\documentclass[11pt]{article}

\usepackage[hyperref]{ccl2023-en}
\usepackage{times}
\usepackage{url}
\usepackage{latexsym}
\usepackage{fancyhdr}

\usepackage{color}
\definecolor{dark-red}{RGB}{254,129,125}
\definecolor{dark-green}{RGB}{129,184,223}

\usepackage{amsmath,amssymb}
\usepackage{tikz}
\usepackage{mathtools}
\DeclarePairedDelimiterX{\infdivx}[2]{(}{)}{%
  #1\;\delimsize\|\;#2%
}
\newcommand{\infdiv}{D_{KL}\infdivx}
\newcommand{\jsdiv}{D_{JS}\infdivx}

\usetikzlibrary{backgrounds}
\usetikzlibrary{fit}
\usepackage{booktabs}
\usepackage{array}
\usepackage{multirow}
\usepackage{pgfplots}
\pgfplotsset{compat=1.18} 
\usepackage{subcaption} 
\usepackage{graphicx}
\title{Towards Robust Aspect-based Sentiment Analysis through Non-counterfactual Augmentations}

\author{Xinyu Liu\textsuperscript{1}, Yan Ding\textsuperscript{1}, \ Kaikai An\textsuperscript{1}, \\ 
\textbf{Chunyang Xiao\textsuperscript{2}, Pranava Madhyastha\textsuperscript{3}, Tong Xiao\textsuperscript{1,4} \ and Jingbo Zhu\textsuperscript{1,4}} \\
	\textsuperscript{1}NLP Lab, Northeastern University, Shenyang, China\\
	\textsuperscript{2}Amazon Alexa\\
	\textsuperscript{3}City, University of London\\
	\textsuperscript{4}NiuTrans Research, Shenyang, China
}

\date{}
\begin{document}
\maketitle 
\begin{abstract}
  While state-of-the-art NLP models have demonstrated excellent performance for aspect based sentiment analysis (ABSA), substantial evidence has been presented on their lack of robustness. This is especially manifested as significant degradation in performance when faced with out-of-distribution data. Recent solutions that rely on counterfactually augmented datasets show promising results, but they are inherently limited because of the lack of access to explicit causal structure. In this paper, we present an alternative approach that relies on non-counterfactual data augmentation. Our proposal instead relies on using noisy, cost-efficient data augmentations that preserve semantics associated with the target aspect. Our approach then relies on modelling invariances between different versions of the data to improve robustness. A comprehensive suite of experiments shows that our proposal significantly improves upon strong pre-trained baselines on both standard and robustness-specific datasets. Our approach further establishes a new state-of-the-art on the ABSA robustness benchmark and transfers well across domains.
\end{abstract}

\section{Introduction}
\label{intro}

%
%

Deep neural models have recently made remarkable advances in aspect-based sentiment analysis (ABSA)~\cite{rietzler2019adapt,xu2019bert,zhang-etal-2019-aspect}. However, recent work has also remarked that a great part of the advances is due to models learning spurious patterns~\cite{DBLP:journals/corr/abs-1711-11561,DBLP:conf/aies/SlackHJSL20}. Consider the sentence 
“Tasty burgers and crispy fries”, for the target word “burgers”, the word that determines the positive sentiment is clearly “tasty” instead of “crispy” or “fries”. Because of the correlation in the dataset, models trained with supervised learning have been shown to be susceptible to associate the positive sentiment of “burgers” with “crispy” or “fries”; We refer to these as spurious patterns in this paper. Spurious patterns partially affect the robustness -- out-of-domain (OOD) generalization of the models trained on independent, identical distribution (IID) data, leading to significant performance decay under distribution shift~\cite{NiallMAdams2010DatasetSI,MasashiSugiyama2012MachineLI,YanivOvadia2019CanYT}. For the task of ABSA, Xing et al. \shortcite{DBLP:conf/emnlp/XingJJWZH20} observed that, current state-of-the-art approaches in ABSA lack robustness and suffer substantially either with out of distribution or adversarial examples, with degradation in the order of over 25\% performance drops. Besides performance, spurious features are also related to issues such as social bias~\cite{RanZmigrod2019CounterfactualDA,RowanHallMaudslay2019ItsAI,DBLP:conf/birthday/LuMWAD20}.

To prevent learning spurious patterns, one of the popular solutions is using counterfactually augmented data (CAD)~\cite{DivyanshKaushik2019LearningTD} to ensure that models learn real causal associations between the input texts and the corresponding labels. For example, a sentiment-flipped counterfactual of the last example could be “Bad burgers but crispy fries”. By this means, we teach the model through data that “crispy fries” don't determine the sentiment for the target aspect and encourage the model to find clues elsewhere. However, generating counterfactual for sentiment analysis is generally hard; as in this example, the generation model is supposed to be able to find the determining causal factor “tasty” at first for the data augmentation process. Such difficulty amplifies in~\cite{PdraigCunningham2022ExploringTE} for implicit expressions.

In this work, we empirically investigate the following research question: \textit{Can we improve model's robustness by using data augmentations other than purely counterfactual data augmentations}? In this paper, we present a method that only relies on automatically-generated noisy yet cost-effective data augmentations that have no dependence with the target aspect in the context of ABSA. By definition, such augmentations will not change the target prediction (Section \ref{method}). Because our augmentation proposal does not require any knowledge of the causal features for the target aspect, such augmentations are significantly cheaper and easier than counterfactual data augmentations. Also, we propose an approach that encourages the model prediction to be invariant between the original sentence and the augmented sentence. When predictions are invariant, the model effectively learns to ignore the spurious features and focus on the information that is core for the target aspect. Our proposal is based on a causal analysis of the ABSA prediction and also presents theoretical guarantees for invariance beyond ABSA.

Furthermore, we show that models trained with our proposal not only avoid spurious patterns that it has seen; through training, the model also learns to generalize over the spurious patterns and avoids to use such information during inference, improving robustness in consequence. We verify this through a comprehensive suite of experiments that includes independent and identically distributed (IID) settings, and out of domain (OOD) settings. We further show the efficacy of the approach in transfer learning settings.

\section{Method}
\label{method}
We first present the theoretical foundations for analysing ABSA through a causal lens, where we also present the theoretical generalisation to other settings. We then present the data augmentation process which is used for the task of ABSA. Finally, we introduce our Causally Relevant Representation (CRR) learning model for leveraging such data.

\subsection{Analysis through causal lens}
\label{section_analysis}
\begin{figure}[t]
\begin{subfigure}[b]{0.4\textwidth}
\begin{tikzpicture}[scale=0.8,>=stealth]
    \tikzstyle{n1} = [draw,minimum size=2em,circle,inner sep=1pt]
    \tikzstyle{n2} = [draw,inner xsep=4pt,minimum height=2em]
    \tikzstyle{t1} =[rectangle,minimum width = 2pt,minimum height=0.5pt]
    \node[n2,fill=gray!50] (x1) at (0,0) {$X$};
    \node[n2,anchor=north] (r1) at ([yshift=-2em]x1.south) {$R(X)$};
    \node[n2,anchor=west] (y1) at ([xshift=4.5em]r1.east) {$Y$};
    \node[n1,anchor=south] (c) at ([xshift=3em,yshift=3em]x1.north)  {$C$};
    \node[n1,anchor=south] (s) at ([xshift=-3em,yshift=3em]x1.north)  {$S$};
    \node[n1,anchor=east] (s1) at ([xshift=-4em,yshift=0.5em]s.west)  {$S_1$};
    \node[anchor=north] at ([yshift=0.3em]s1.south)  {$\vdots$};
    \node[n1,anchor=east] (sn) at ([xshift=-4em,yshift=-4.5em]s.west)  {$S_n$};
    \begin{pgfonlayer}{background}
    \node[draw,rounded corners=2pt,inner sep=8pt,label=above:\small{Data composition}][fit = (x1)(c)(s)(s1)(sn)]{};
    \node[draw,rounded corners=2pt,inner sep=8pt,label=below:\small{Learning process}][fit = (x1)(r1)(y1)]{};

    \end{pgfonlayer}
    \draw[-latex] (s1.0) -- (s.180);
    \draw[-latex] (sn.0) -- (s.180) ;
    \draw[-latex] (c.-90) -- (x1.90);
    \draw[-latex] (s.-90) -- (x1.90);
    \draw[-latex,out=-10,in=80] (c.-30) to (y1.90);
    \draw[-latex,dashed] (x1.-90) -- (r1.90);
    \draw[-latex,dashed] (r1.0) -- (y1.180);
    \end{tikzpicture}
    \caption{\label{fig:causal}}
\end{subfigure}
\quad
\quad
\begin{subfigure}[b]{0.5\textwidth}
\includegraphics[width=1\textwidth]{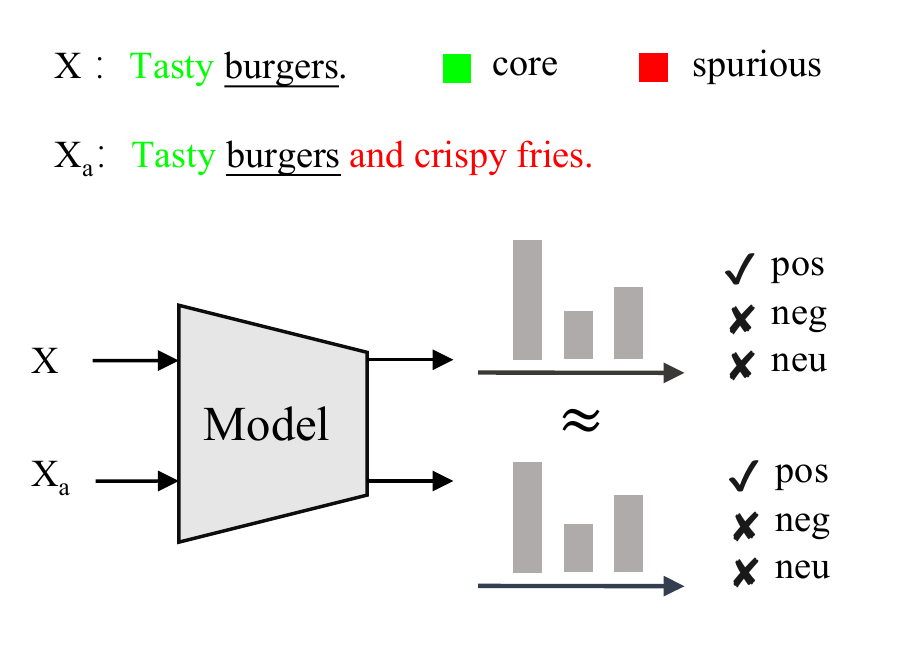}
\caption{\label{fig:invarient}}
\end{subfigure}
\caption{\textbf{(a)} Causal graph formalizing assumptions about core ($C$) and spurious ($S$) features of the data ($X$) and the relationship between the features and ABSA target. \textbf{(b)} An instance to distinguish $C$ and $S$ features for the \underline{target aspect}, our model shows invariance on predictions when appending $S$.}
\end{figure}
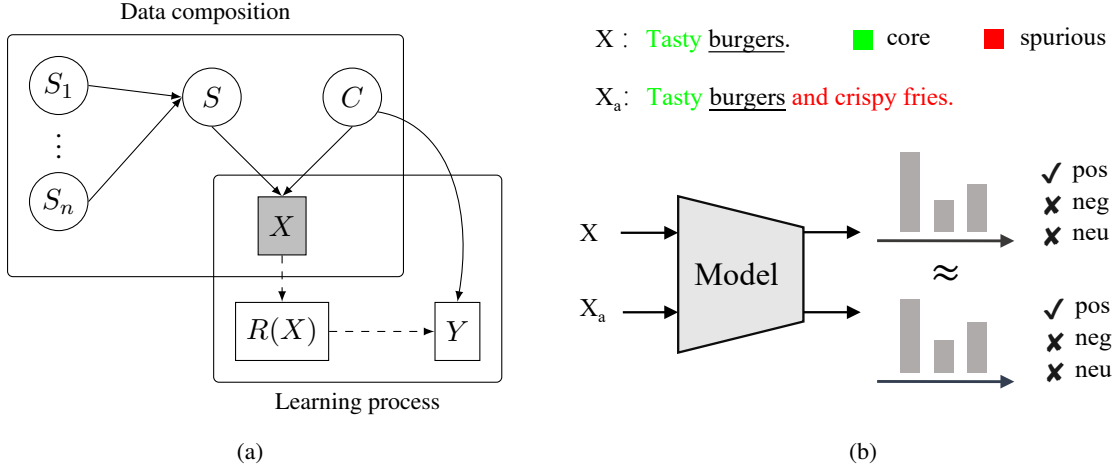

To formally distinguish between core and spurious features and to perform analysis over our ABSA settings, we propose to analyse ABSA through the causal lens. We specifically intend to distinguish two types of features: a) \emph{Core features ($C$)} are features that causally determine the label for a given sample; and b) \emph{Spurious features ($S$)} are features that might contribute to the correct prediction because of dataset biases. While statistically, they may both be associated with the label with different predictive strengths, a causal analysis allows us to have a different perspective. In Figure~\ref{fig:causal}, we draw the causal graph for our setup.\footnote{Causal graph is a directed graph where the arrows from causes to effects.} Through this causal view, we see that both core structures $C$ and spurious structures $S$ form the sentence $X$, however, only $C$ determines the label $Y$. In other words, in this causal setup, the sentiment of label $Y$ can be changed only by significantly changing $C$. To illustrate with an example, consider the sentence “tasty burgers and crispy fries”, for the aspect “burgers”, $C$ contains the most influential word “tasty” while all the other words in the sentence (i.e. “and”, “crispy”, “fries”) are spurious features and belong to $S$ to analyse the sentiment for “burgers”. According to Figure~\ref{fig:causal}, $S$ should not interfere the model's predictions; we show such invariance in Figure~\ref{fig:invarient}.

Because $C$ has all the necessary information to predict $Y$ and $S$ is causally independent of $C$, the independence mechanisms~\cite{JonasPeters2017ElementsOC} show that, for the input data $X=f(C,S)$, any interventions on $S$ should not change the conditional distribution $P(Y\,\vert\,X)$, i.e.
\begin{equation}
p(Y\,\vert\,X)=p(Y\,\vert\,do(S\!=\!s),X)\quad\forall s\in\mathcal{S}.\notag
\label{invariant_eq}
\end{equation}
$\mathcal{S}$ is the domain of spurious feature $S$ and $do(S\!=\!s)$ denotes assigning $S$ the value $s$~\cite{pearl2009causality}. We further operationalize this equation by learning \emph{invariant representation $R(X)$} under data augmentations, which must fullfill the following criteria:
\begin{equation}
p(Y\,\vert\,R(X))=p(Y\,\vert\,do(s_i),R(X))\quad\forall s_i\in\mathcal{S},
\end{equation}
where $\mathcal{S}=\{s_i\}_{i=1}^N$ is the set of augmentations which intervene the spurious features $S$. For example, the \textsc{AddDiffMix} augmented samples in Table~\ref{tab:ins_compare} change $S$ by appending contents which are irrelevant to the target aspect.

The equations above depict the behaviour of an ideal model: when making changes on spurious patterns, the model should not change its prediction of the corresponding label. A model that has the above properties must show robustness to changes in spurious information for all tasks that follow the causal graph shown in Figure~\ref{fig:causal}. We formalise this in the following theorem\footnote{Please refer to Appendix~\ref{appendix:causal theorem} for the accompanying proofs.}: 

\setlength{\parskip}{5pt}
\noindent
\textbf{Theorem.} \emph{Let $\mathcal{Y}=\{Y_t\}_{t=1}^T$ be a family of tasks, and $Y^R$ be a specific task cover all the causal relationships in $\mathcal{Y}$. If $R(X)$ is an invariant representation for $Y^R$ under spurious pattern interventions $S$, then $R(X)$ is an invariant representation for all tasks in $\mathcal{Y}$ under spurious pattern interventions $S$, i.e.
\begin{equation}
\begin{aligned}
p(Y^R\,\vert\,R(X))&=p(Y^R\,\vert\,do(s_i),R(X))\Rightarrow
\\
p(Y_t\,\vert\,R(X))&=p(Y_t\,\vert\,do(s_i),R(X))\quad\forall s_i\in\mathcal{S},Y_t\in\mathcal{Y}\notag
\label{theorem}
\end{aligned}
\end{equation}
Thus $R(X)$ is a representation generalized to $\mathcal{Y}$.}

Another important remark for Equation~\ref{invariant_eq} is that although we rely on the core features to prove the invariance, the equation itself does not involve any knowledge about the core features. This implies that we could rely on non counterfactual data augmentations, and leverage the invariance in Equation~\ref{invariant_eq} to improve model robustness.
\setlength{\parskip}{0pt}

\subsection {Data augmentation for ARTs Dataset}
\label{Data Aug}
In this work, we are especially interested in questions relating to \emph{aspect robustness} in ABSA models. To investigate the aspect robustness of ABSA models, Xing et al. \shortcite{DBLP:conf/emnlp/XingJJWZH20} proposes an aspect robustness test set (ARTs) that extends the test sets in SemEval2014 datasets~\cite{pontiki-etal-2014-semeval} with three data augmentation operations: (1) \textbf{\textsc{RevTgt}} augments the samples with tokens that reverse the sentiment of the target aspect. (2) \textbf{\textsc{RevNon}} augments the samples with tokens which retain the sentiment of the target aspect, but change sentiments of all non-target aspects. (3) \textbf{\textsc{AddDiff}} appends the samples with new segments involving aspects different from the target aspect.\footnote{\textsc{RevTgt} and \textsc{RevNon} could only apply to the instances with explicit opinion words, while \textsc{AddDiff} could operate on all the instances.} These modifications are based on a set of hand crafted heuristics. Using the causal approach introduced in Section \ref{section_analysis}, we can see the \textbf{\textsc{RevTgt}} operation making changes to the core feature $C$, and thus consist of counterfactual augmentations. The \textbf{\textsc{RevNon}} and \textbf{\textsc{AddDiff}} operations on the other hand, don't influence the target prediction and only operate on $S$. In this paper, we build on this observation and choose a cost-effective \textbf{\textsc{AddDiff}} that doesn't rely on any access to causal features unlike counterfactual data augmentation methods.

\begin{table}[t]\footnotesize
\centering
\resizebox{0.98\textwidth}{!}{
\begin{tabular}{p{3cm}p{11cm}}
\toprule[1pt]
    \multicolumn{1}{c}{\textbf{Scenario}} & \multicolumn{1}{c}{\textbf{Instance}} \\
\midrule[0.5pt]
    Original & \underline{3D rendering} slows it down considerably.\\
\midrule[0.5pt]
    Standard \textsc{AddDiff} & \underline{3D rendering} slows it down considerably, \textcolor{red}{but \underline{keyboard} is a love, \underline{battery life} is amazing and \underline{quality} is a superlative.}\\
\midrule[0.5pt]
    \textsc{AddDiff}-\textsc{Mix} (Front) & \textcolor{red}{Although the very knowledgeable \underline{sales associate} and \underline{screen size} is perfect for portable use in any environment,} \underline{3D rendering} slows it down considerably.\\
\midrule[0.5pt]
    \textsc{AddDiff}-\textsc{Mix} (Rear) & \underline{3D rendering} slows it down considerably, \textcolor{red}{but for the \underline{price}, I was very pleased with the condition and the overall product and my new Toshiba \underline{works} great on both.}\\
\bottomrule[1pt]
\end{tabular}}
\caption{A sample from the SemEval 14 Laptop testset, its \textsc{AddDiff} + manual revision counterpart from ARTs and the samples generated by our reimplementation of \textsc{AddDiff}, i.e. \textsc{AddDiffMix}. Augmented context is indicated in red and aspects are underlined.}
\label{tab:ins_compare}
\end{table}

Table~\ref{tab:ins_compare} illustrates how our augmentation method \textsc{AddDiffMix} operates over the original sentence to generate data. More specifically, our method begins by constructing an $\rm AspectSet$ by extracting all aspects together with their sentiment expressions from the entire dataset using pretrained constituency parsing~\cite{DBLP:conf/acl/HopkinsJP18}; for example, we extract the item “Food at a reasonable price” with positive sentiment polarity for the aspect “Food”. Using the $\rm AspectSet$, \textsc{AddDiffMix} operation then randomly samples 1-3 phrases which are not mentioned in the original sample and whose sentiments are opposite to the target aspect's sentiment. We then prepend or append those over the original sample with a 50/50 split. We note that, while our method is inspired by the original \textsc{AddDiff} method, they differs in two ways. Firstly, as shown in Table~\ref{tab:ins_compare}, we consider both prepend and append operations over the original sentence while the original \textsc{AddDiff} only considers an append operation. Secondly, the original \textsc{AddDiff} method leverages manually crafted heuristics to further correct the sentence for fluency while we do not, making our method more generally applicable.

\subsection{The CRR Model}
\label{CRR Method}
We present a general method explicitly presumes that the data augmentations under consideration only introduce spurious information. The main idea is to promote the predicted probabilities of the original input and the corresponding augmented input to be close to each other following Equation~\ref{invariant_eq}. Through learning to ignore the spurious information, we implicitly encourage the model to focus more on causally relevant parts and generate causally relevant representations (CRR) for the prediction of the correct label.

Formally, our approach is an instance level method; for each input instance and its corresponding label $x,y$ from the training dataset $\mathcal{D}$, let $P(y\,\vert\,x)$ be the conditional probability of the original data, and $P(y\,\vert\,x^a)$ be the conditional probability distribution of the corresponding augmented data where $x^a$ denotes the input after applying the data augmentation that introduces only spurious features over $x$. As the standard approach in data augmentation literature, we minimize the cross entropy loss over both the original and augmented data: 
\begin{equation}
\mathcal{L}_{CE}=-\log\,p(y\,\vert\,x)-\log\,p(y\,\vert\,x^a)\quad\forall x,y\in\mathcal{D}
\end{equation}
As we consider data augmentation process that a priori only introduces spurious features, the probability remains invariant following Equation~\ref{invariant_eq}. We introduce a regularization term based on the Kullback-Leibler (KL) divergence to favor model output probabilities (i.e. the prediction over the original data and the prediction over the corresponding augmented data) to be close to each other.\footnote{We have also investigated the KL divergence from the other direction (as the KL-divergence is asymmetric) and the Jenshen Shannon (JS) divergence which give similar results as our proposed method, these are presented in the Appendix (Table~\ref{tab:divergence}).}  
\begin{equation}
\mathcal{L}_{KL}=\infdiv{p(y\,\vert\,x)}{p(y\,\vert\,x^a)}\quad\forall x,y\in\mathcal{D}
\label{kl}
\end{equation}
This loss intends to minimise the difference between the two probability distributions calculated over the original sentence and the augmented sentence.
We then perform a weighted sum of the two loss functions: 
\begin{equation}
\mathcal{L}_{CRR}=\sum_{\mathcal{D}}(\mathcal{L}_{CE}+\alpha\mathcal{L}_{KL})
\label{objective}
\end{equation}
where, $\alpha$ is a scalar to control the extent of regularisation. 

\section{Experiments}
\subsection{Data and Processing}
We conduct experiments on the SemEval2014 Laptop and Restaurant Reviews datasets (Laptop and Restaurant) \cite{pontiki-etal-2014-semeval} and their ARTs~\cite{DBLP:conf/emnlp/XingJJWZH20} extensions. We follow the previous studies and remove instances with conflicting polarity \cite{wang2016attention,ma2017interactive,xu2019bert} and use the train-dev split as stated in \cite{DBLP:conf/naacl/XuLSY19}. As detailed in Section~\ref{Data Aug}, for the CRR models, we perform data augmentation by applying \textsc{AddDiffMix} to the training sets from Laptop and Restaurant; we apply \textsc{RevTgt} to compare with counterfactual generation settings.

\subsection{Models}
\label{models}
Previous works have observed that algorithms that exploit pretrained models exhibit more robustness ~\cite{DBLP:journals/corr/abs-2103-00020,DBLP:journals/corr/abs-2004-06100} for tasks including ABSA~\cite{DBLP:conf/emnlp/XingJJWZH20}. Inspired by this, we also focus on using pretrained models in our paper, making use of the same RoBERTa-large based model RoBERTa$_\textrm{L}$ as in Dai et al. \shortcite{DBLP:conf/naacl/DaiYSLQ21}. The model adds a multilayer perceptron on top of the pooled features induced with RoBERTa$_\textrm{L}$ \cite{DBLP:journals/corr/abs-1907-11692} and is then finetuned on the original SemEval2014 data. We use this as our baseline for our paper. We finetune this model on the combined dataset consisting of original and counterfactually-augmented data to establish the counterfactual baseline. We also compare with the following models in the literature: 

a) \textbf{BERT-PT}: Xu et al.~\shortcite{xu2019bert} developed a BERT based model which is first trained on review datasets such as Amazon laptop reviews \cite{DBLP:conf/www/HeM16} and then is finetuned on the ABSA task. It is reported to be the best performing model in \cite{DBLP:conf/emnlp/XingJJWZH20}; 

b) \textbf{GraphMerge}: Hou et al.~\shortcite{DBLP:conf/naacl/HouQWYHHZ21} combine the dependency relations from different parses and applies RGAT \cite{DBLP:journals/corr/abs-1904-05811} over the resulting graph. The proposed method is reported to achieve SOTA results on ABSA, as well as, be able to improve on robustness measured by ARTs;

c) \textbf{RoBERTa+POS bias}: Ma et al.~\shortcite{DBLP:conf/acl/MaZS21} inject position-bias weight into fine tuning RoBERTa model. The proposed method mainly improves the model robustness measured by ARTs.

d) \textbf{BERTAsp+SCAPT}: Li et al.~\shortcite{DBLP:conf/emnlp/LiZZZW21} applies supervised contrastive pre-training training (SCAPT), as well as, continuing to pretrain BERT over large scale external in-domain data, and then fine tunes the model on the SemEval2014 datasets. It is reported to improve on both the original dataset and ARTs.

\subsection{Implementation Details and Metrics}
We fine tune all the pretrained models with a batch size $b = 64$, a dropout rate $d = 0.3$, and an AdamW optimizer~\cite{DBLP:conf/iclr/LoshchilovH19} with warm-up for both the \textit{Laptop} and  \textit{Restaurant} datasets. We perform a grid search over learning rates $\{5e^{-6}, 1e^{-5}, 2e^{-5}\}$ for both datasets in all experiments; for CRR that we propose, we also grid search over the regularization weights $\{1, 3, 5\}$.
We train all the models up to 40 epochs and select the best model according to the result on the validation set, which we set to the original validation set.~\footnote{We are aware of the limitations of such choices as pointed out in \cite{DBLP:conf/emnlp/CsordasIS21}; however, given that our objective is to generalize to all \textit{unknown} O.O.D settings, we consider the original validation set a sensible choice.} For comparison, we report the accuracy, \textbf{aspect robustness scores (ARS)} and macro F1 scores that are averaged over 5 experiments. \textbf{ARS} is a robustness focused metric introduced in~\cite{DBLP:conf/emnlp/XingJJWZH20} where a model is considered to have answered correctly a question only if all variations of that question (i.e. its \textsc{AddDiff}, \textsc{RevTgt}, \textsc{RevNon} variants) are answered correctly. All the implementations are based on fastNLP.

\section{Results and Analysis}
\label{results}

\subsection{Main Results}
The results are shown in Table~\ref{tab:main_result} and detailed in the following parts.

\begin{table*}[h]\footnotesize
\centering 
\resizebox{0.98\textwidth}{!}{
\begin{tabular}{lcccccccccc}
\toprule[1pt]
\multirow{3}{*}{Model} 
  & 
  \multicolumn{5}{c}{\textsc{Laptop}} & \multicolumn{5}{c}{\textsc{Restaurant}}\\
  \cmidrule(lr){2-6}\cmidrule(lr){7-11} 
  &
  \multicolumn{2}{c}{Original} & \multicolumn{3}{c}{ARTs} & \multicolumn{2}{c}{Original} & \multicolumn{3}{c}{ARTs} \\ 
  \cmidrule(lr){2-3}\cmidrule(lr){4-6}\cmidrule(lr){7-8}\cmidrule(lr){9-11}
  &
  F1 & Acc. & F1 & Acc. & ARS & F1 & Acc. & F1 & Acc. & ARS\\
\midrule[0.5pt]
BERT-PT
& -- & 78.53 & -- & 73.37$\dagger$ & 53.29 & -- & 86.70 & -- & 80.71$\dagger$ & 59.29 \\
GraphMerge
& -- & 80.00 & -- & -- & 52.90 & -- & 85.16 & -- & -- & 57.46 \\
RoBERTa+POS bias
& -- & -- & 72.09 & 75.72 & -- & -- & -- & 73.10 & 79.47 & -- \\
BERTAsp+SCAPT
& 79.15 & 82.76 & 72.80* & 76.13* & 57.84* & \textbf{83.79} & \textbf{89.11} & 74.22* & 80.03* & 56.43* \\
\midrule[0.5pt]
RoBERTa${\rm_L}$
& 79.23 & 82.63 & 73.90 & 77.32 & 59.06 & 79.11 & 86.73 & 74.62 & 81.33 & 59.48 \\
RoBERTa${\rm_L}$+CAD
& 77.47 & 81.35 & 72.83 & 76.66 & 61.55 & 79.25 & 85.80 & \textbf{76.64} & 81.81 & \textbf{64.29} \\
RoBERTa${\rm_L}$+CRR
& \textbf{80.65} & \textbf{83.86} & \textbf{76.62} & \textbf{79.66} & \textbf{64.58} & 80.85 & 87.63 & 76.60 & \textbf{82.79} & 63.30\\
\bottomrule[1pt]
\end{tabular}}
\caption{Accuracy, F1 and ARS results(\%) on the Laptop and the Restaurant datasets. \textbf{Original} setting tests on the SemEval2014 test set and \textbf{ARTs} setting tests on its ARTs counterpart. We report accuracy and macro-F1 for \textbf{Original} and \textbf{ARTs} and ARS in addition for \textbf{ARTs}. Texts in bold indicate the best results. * denotes the results of our reruning version, $\dagger$ denotes the results that we calculate based on the reported scores.}
\label{tab:main_result}
\end{table*}

\paragraph{Performance of Baselines}
Our implementation of the RoBERTa$_\textrm{L}$ baseline outperforms the reported methods on ARTs and also has competitive performance on the original data. Specifically, on ARTs Laptop/Restaurant, RoBERTa$_\textrm{L}$ outperforms the best models in the literature by 1.22\% and 0.19\% in ARS respectively. On the original test sets, RoBERTa$_\textrm{L}$ performs best amongst models without external data, which improves significantly compared to weaker pretrained models BERT and RoBERTa-base. Our study illustrates again the potentials of foundation models~\cite{DBLP:journals/corr/abs-2108-07258} to improve i.i.d performance and robustness. 

\paragraph{Performance of Counterfactual Models} 
Compared with the RoBERTa$_\textrm{L}$ baseline, the counterfactual model, i.e. RoBERTa$_\textrm{L}$+CAD (coutefactually augmented data), has limited improvement (0.48\% in accuracy) in Restaurant ARTs, even suffers performance degradation (-0.66\% in accuracy) in Laptop ARTs. What's more, the CAD model degrades the performance on the original Restaurant/Laptop test sets. Similar findings were reported by Cunningham et al. \shortcite{PdraigCunningham2022ExploringTE}, where fine-tuning on the combined dataset consisting of original and the CAD datasets sometimes affects the model in-domain performance.
 
\paragraph{Performance of CRR Models}
Our proposed CRR model (RoBERTa$_\textrm{L}$+CRR) shows significant improvements in ARTs on both the Laptop and the Restaurant datasets, which outperforms the strong RoBERTa$_\textrm{L}$ baseline by 2.34\% in accuracy (5.52\% in ARS) for the Laptop dataset and by 1.46\% in accuracy (3.82\% in ARS) for the Restaurant dataset respectively. Our model also improves on the original datasets, thus bringing improvements over all testing cases. These results suggest that our approach is particularly promising in improving general robustness since robustness focuses on all potentially encountered distributions. 

\subsection{Fine grained analysis}
\label{ModelAnalysis}

\begin{table}[t]
\centering
\resizebox{0.65\textwidth}{!}{
\begin{tabular}{clcclccl}
\toprule[1pt]
\multirow{2}{*}{Test Set} & \multicolumn{1}{c}{\multirow{2}{*}{Model}} & \multicolumn{3}{c}{Laptop} & \multicolumn{3}{c}{Restaurant} \\ \cmidrule(l){3-5}\cmidrule(l){6-8}
                          & \multicolumn{1}{c}{} & Original & ARTs & \multicolumn{1}{c}{Diff} & Original & ARTs & \multicolumn{1}{c}{Diff} \\
\midrule[0.5pt]
\multirow{2}{*}{RevTgt} & baseline & 84.21 & 61.46 & \textcolor{red}{$\downarrow$22.75} & 92.84 & 64.04 & \textcolor{red}{$\downarrow$28.80} \\
                         & CAD      & 82.32 & 74.16 & \textcolor{red}{$\downarrow$8.16} & 91.42 & 82.27 & \textcolor{red}{$\downarrow$9.15} \\
                         & CRR      & 84.68 & 64.42 & \textcolor{red}{$\downarrow$20.26} & 92.98 & 65.72 & \textcolor{red}{$\downarrow$27.26} \\
\midrule[0.5pt]
\multirow{2}{*}{RevNon} & baseline & 94.52 & 83.70 & \textcolor{red}{$\downarrow$10.82} & 93.87 & 84.82 & \textcolor{red}{$\downarrow$9.05} \\
                         & CAD      & 90.96 & 55.11 & \textcolor{red}{$\downarrow$35.85} & 92.84 & 69.91 & \textcolor{red}{$\downarrow$22.93} \\
                         & CRR      & 93.33 & 85.78 & \textcolor{red}{$\downarrow$7.55} & 93.65 & 86.04 & \textcolor{red}{$\downarrow$7.61} \\
\midrule[0.5pt]
\multirow{2}{*}{AddDiff} & baseline & 82.63 & 80.47 & \textcolor{red}{$\downarrow$2.16} & 86.73 & 87.46 & \textcolor{green}{$\uparrow$0.73} \\
                         & CAD      & 81.35 & 66.55 & \textcolor{red}{$\downarrow$14.80} & 85.80 & 82.11 & \textcolor{red}{$\downarrow$3.69} \\
                         & CRR      & 83.86 & 84.64 & \textcolor{green}{$\uparrow$0.78} & 87.63 & 89.30 & \textcolor{green}{$\uparrow$1.67} \\
\bottomrule[1pt]
\end{tabular}}
\caption{The model accuracy(\%) on \textbf{Original} subsets as well as the corresponding \textbf{ARTs} version where the generation strategies \textsc{RevTgt}, \textsc{RevNon} and \textsc{AddDiff} are applied. We also report the \textbf{Diff}erence of models' performance on these two sets. \textcolor{red} {Red} indicates decline while \textcolor{green} {green} indicates rising.}
\label{tab:ana_adiff}
\end{table}

\paragraph{How does CRR, CAD and baseline model behave on ARTs's subsets?}
We want to investigate the performance gains on ARTs from our CRR proposal and make a fine-grained comparison with the performance of the CAD models. As elucidated in Section \ref{Data Aug}, ARTs augments the original test set using three operations (i.e. \textsc{RevTgt}, \textsc{RevNon} and \textsc{AddDiff}). We here analyse \underline{a)} if the improvement of our models we observe in Table~\ref{tab:main_result} is predominantly due to one particular domain (i.e the data corresponding to the \textsc{AddDiff} operation), as the augmentation for our method only exploits operations similar to the \textsc{AddDiff} operations, refer Section \ref{Data Aug}). \underline{b)} the root cause of the mediocre performance of CAD models.

a) Our results are shown in Table~\ref{tab:ana_adiff}. We observe that the CRR model improves over baseline not only over the {ARTs}, \textsc{AddDiff} cases but also in all the other cases involving original datasets and other operations. Notably, we observe that for non \textsc{AddDiff} cases, while our proposed CRR model may still suffer from degradation, it always degrades less compared to the baselines. For example, for Laptop, CRR degrades by 20.26\% and 7.55\% on \textsc{RevTgt} and \textsc{RevNon} respectively, compared to 22.75\% and 10.82\% for the baseline. These results suggest that our proposed approach indeed helps the model to focus on core predictive features so that it improves not only on its trained data distribution but also on other distributions. We also refer the readers to Appendix~\ref{appendix:AddDiff} for more details on experiments comparing the \textsc{AddDiff} and  \textsc{AddDiffMix} operations. 

b) As shown in Table~\ref{tab:ana_adiff}, the CAD models significantly improve the performance over the {ARTs}, \textsc{RevTgt} cases but perform worst on all the other Original/New subsets, which is consistent with the overall mediocre results seen in Table~\ref{tab:main_result}. The results suggest that although the CAD method could be effective on the specific data aligned with counterfactual augmentation, it might harm model performance measured on other data.

\paragraph{How does our proposed method compare to adversarial training?}
As shown in Xing et al.~\shortcite{DBLP:conf/emnlp/XingJJWZH20}, a simple way to leverage augmented data is to perform adversarial training on such data. We are interested in whether such a simple process can also improve the general model performance on the original data and ARTs. Specifically, we first inject our newly generated \textsc{AddDiffMix} data into the original train data  and then fine tune the pre-trained models on it. The adversarial models share the same hyperparameters setting with baselines and CRR models. The results are presented in Table~\ref{tab:adversarial}. By comparing the performance of RoBERTa${\rm_L}$ and the adversarial training on top of that, we can see that adversarial training fails to consistently improve model performance. 

Specifically, the adversarial models (+Adv) have very limited accuracy improvements on the original and ARTs datasets; and the ARS score is even worse than that of the RoBERTa${\rm_L}$ baseline. We believe that this is because our augmented data consists solely of~\textsc{AddDiffMix} data and doesn't cover, for example, data generated from other operations. This is in contrast to Xing et al.~\shortcite{DBLP:conf/emnlp/XingJJWZH20}'s finding where data augmentations come from all three operations. This result shows that in real-life scenarios where data augmentation doesn't align well with the test distribution, simply training on augmented data has very limited effect on improving the general performance. From the table, we also see that CRR performs significantly better than adversarial training with the same training data. We believe that via the invariance loss (i.e. Equation~\ref{kl}) incorporated into the model training process, the resulting model learns more to focus on the predictive features and thus can improve performance on all the datasets as shown in Table~\ref{tab:ana_adiff} and Table~\ref{tab:adversarial}.

\begin{table}[t]
\centering
\resizebox{0.4\textwidth}{!}{
\begin{tabular}{lccc}
\toprule[1pt]
\multirow{2}{*}{Method} 
  & 
  \multicolumn{3}{c}{\textsc{Restaurant}} \\
  \cmidrule(lr){2-4}
  &
  Acc.(O) & Acc.(N) & ARS(N) \\
\midrule[0.5pt]
BERT
& 83.02 & 75.37 & 50.59 \\
\ \ (+Adv)
& 83.59 & 79.92 & 57.17 \\
\ \ (+CRR)
& 83.70 & 80.14 & 59.23 \\

RoBERTa
& 86.25 & 79.44 & 56.30 \\
\ \ (+Adv)
& 85.74 & 80.86 & 57.26 \\
\ \ (+CRR)
& 86.45 & 81.89 & 62.29 \\

RoBERTa${\rm_L}$
& 86.73 & 81.33 & 59.48 \\
\ \ (+Adv)
& 86.86 & 81.95 & 59.27 \\
\ \ (+CRR)
& \textbf{87.63} & \textbf{82.79} & \textbf{63.30}\\
\bottomrule[1pt]
\end{tabular}}
\caption{The accuracy of different models on Restaurant (Acc.(O)) and the corresponding ARTs (Acc.(N)); for ARTs, we also show ARS (ARS(N)). The models vary on pretrained settings and training strategies. }
\label{tab:adversarial}
\end{table}

\paragraph{Do our results generalize to other pretrained models?}
To test if the improvement we observe with CRR over baseline and adversarial training is specific to the RoBERTa${\rm_L}$ pretrained model, we experiment with two other pretrained models for which we show results in Table~\ref{tab:adversarial}. We see that adversarial training helps somewhat when pretrained model is weaker (i.e BERT); nevertheless, our proposed model consistently outperforms both the baseline and adversarial training across all experimental settings.

\subsection{Transfer Ability}
\label{transferability}
Our primary aim is to obtain models that focus on core features and are able to ignore spurious ones. If this is true, empirically, the model should be able to achieve good performance on other ABSA datasets, as the core features are those features that are stable and transfer across datasets, formalized in the theorem in section~\ref{section_analysis}. To test the hypothesis on whether the CRR model actually learns core ABSA features, we perform experiments over Multi Aspect Multi-Sentiment (MAMS) \cite{DBLP:conf/emnlp/JiangCXAY19}. MAMS and its ARTs version (MAMS-ARTs) are challenging datasets for aspect-based sentiment analysis with more complex sentences involving generally multiple aspects. For experiments, we first fine tune the RoBERTa${\rm_L}$ and RoBERTa${\rm_L}$+CRR models respectively on the SemEval2014 Restaurant training set. We then fine tune them by sampling only 10\% of the original MAMS training data and select models based on MAMS's original dev set. The results are presented in Table~\ref{tab:transfer}. By comparing our Roberta based model with and without CRR training, we see that the CRR version improves over all metrics in both MAMS and MAMS-ARTs. This suggests that the CRR model is more robust to distribution changes and has better transfer ability across datasets. Our model uses only 10\% MAMS training data and already performs better than some dedicated models, such as CapsNet~\cite{DBLP:conf/emnlp/JiangCXAY19}, although it is still behind SOTA result BERT-PT.

\begin{table}[h] 
\centering
\resizebox{0.6\textwidth}{!}{
\begin{tabular}{lccccc}
\toprule[1pt]
\multirow{2}{*}{Model} & 
\multicolumn{2}{c}{MAMS} &
\multicolumn{3}{c}{MAMS-ARTs} \\ \cmidrule(l){2-3}\cmidrule(l){4-6}
& F1 & Acc. & F1 & Acc. & ARS \\
\midrule[0.5pt]
BERT-PT & -- & 85.10 & -- & -- & 64.37 \\

CapsNet & -- & 76.05 & -- & 66.09 & 50.90 \\
\midrule[0.5pt]

\textsc{RoBERTa-Trans} & 79.21 & 79.60 & 71.62 & 71.65 & 55.64 \\

\textsc{RoBERTa-Trans-CRR} & 80.14 & 80.47 & 73.24 & 73.22 & 57.24 \\

\bottomrule[1pt]
\end{tabular}}
\caption{The model performance on MAMS test set and its ARTs version.}
\label{tab:transfer}
\end{table}

\subsection{Case Study}

\begin{table*}[t]
\centering
\resizebox{0.98\textwidth}{!}{
\begin{tabular}{p{10cm}p{2cm}p{2cm}p{2cm}}
\toprule[1pt]
    \textbf{Sample Sentences} & \multicolumn{1}{c}{\textbf{RoBERTa}} & \multicolumn{1}{c}{\textbf{Adv}} & \multicolumn{1}{c}{\textbf{CRR}} \\
\midrule[0.5pt]
    1) Would you ever believe that when you complain about over an hour \textbf{\textit{wait}}$_\textbf{\textcolor{green}{N}}$, when they tell you it will be 20-30 minutes, the \textbf{\textit{manager}}$_\textbf{\textcolor{green}{N}}$ tells the \textbf{\textit{bartender}}$_\textbf{\textcolor{green}{O}}$ to spill the \textbf{\textit{drinks}}$_\textbf{\textcolor{green}{O}}$ you just paid for? & N, N, \textcolor{red}{N}, O & N, N, O, O & N, N, O, O \\ 
\midrule[0.5pt]
    2) Frankly, the \textbf{\textit{chinese food}}$_\textbf{\textcolor{green}{N}}$ here is something I can make better at home. & \textcolor{red}{P} & \textcolor{red}{P} & N \\
\midrule[0.5pt]
    3) Great \textbf{\textit{food}}$_\textbf{\textcolor{green}{P}}$ but the \textbf{\textit{service}}$_\textbf{\textcolor{green}{N}}$ was dreadful! & P, N & P, N & P, N \\ 
\midrule[0.5pt]
    4) This place has the strangest \textbf{\textit{menu}}$_\textbf{\textcolor{green}{N}}$ and the restaurants tries too hard to make fancy \textbf{\textit{food}}$_\textbf{\textcolor{green}{N}}$. & N, \textcolor{red}{O} & N, \textcolor{red}{O} & N, N \\ 
\midrule[0.5pt]
    5) I came to fresh expecting a great \textbf{\textit{meal}}$_\textbf{\textcolor{green}{N}}$, and all I got was marginally so-so \textbf{\textit{food}}$_\textbf{\textcolor{green}{N}}$ \textbf{\textit{served}}$_\textbf{\textcolor{green}{O}}$ in a restaurant that was just so freezing we couldn't enjoy eating. & N, N, \textcolor{red}{N} & \textcolor{red}{P}, N, \textcolor{red}{N} & N, N, \textcolor{red}{N} \\ 
\bottomrule[1pt]
\end{tabular}}
\caption{Examples from testset and the predictions of RoBERTa${\rm_L}$, RoBERTa${\rm_L}$+Adv and RoBERTa${\rm_L}$+CRR respectively. The \textbf{\textit{aspects}} are annotated with \textcolor{green}{true labels} in subscripts where P, N, and O denote positive, negative, and neutral respectively. We highlight incorrect predictions in \textcolor{red}{red} color.}
\label{tab:case_study_ori}
\end{table*}

We conduct case studies for the RoBERTa${\rm_L}$, RoBERTa${\rm_L}$+Adv (i.e adversarial training) and RoBERTa${\rm_L}$+CRR models respectively on Restaurant dataset and show some representative cases in table~\ref{tab:case_study_ori}.\footnote{The dev set of the original ABSA dataset contains only 150 instances, so we choose the test set (1,120 instances) to conduct the case studies.} We see that for easy cases where the aspects' sentiments are determined by corresponding explicit opinion word(s) such as in sentence 3), all our models can classify correctly in most test cases. CRR notably improves for long and complex sentences with multiple aspects, as illustrated by sentence 1), 2), 4).
As shown by other experiments, we believe that such improvement can be explained by CRR being able to focus on the core predictive features. We also notice that CRR still struggles with aspects whose sentiments are determined by implicit descriptions, as shown in sentence 5). Such cases are typically hard to learn and one might need extra data to tackle the issue~\cite{li2021learning}.

\begin{table}[t]\footnotesize
\centering
\resizebox{0.98\textwidth}{!}{
\begin{tabular}{p{2cm}p{12cm}p{2cm}}
\toprule[1pt]
\textbf{Model} & \textbf{Example} & \textbf{Pred./Label} \\ 
\midrule[0.5pt]
baseline & \underline{3d rendering} \colorbox[RGB]{250,240,230}{not} \colorbox[RGB]{239,166,102}{slows} it \colorbox[RGB]{239,166,102}{down} \colorbox[RGB]{239,166,102}{considerably}. & neg./pos. \\ 
\midrule[0.5pt]
CRR & \underline{3d rendering} \colorbox[RGB]{239,166,102}{not} \colorbox[RGB]{239,166,102}{slows} \colorbox[RGB]{239,166,102}{it} \colorbox[RGB]{255,235,205}{down} \colorbox[RGB]{255,235,205}{considerably}. & pos./pos. \\
\midrule[0.5pt]
baseline & Great \underline{food}, \colorbox[RGB]{239,166,102}{not} \colorbox[RGB]{239,166,102}{great} \colorbox[RGB]{239,166,102}{waitstaff}, \colorbox[RGB]{255,235,205}{not} great \colorbox[RGB]{255,235,205}{atmosphere}, but worst of all not GREAT beer! & neg./pos. \\
\midrule[0.5pt]
CRR & Great \underline{food}, \colorbox[RGB]{239,166,102}{not} \colorbox[RGB]{239,166,102}{great} waitstaff, \colorbox[RGB]{255,235,205}{not} great atmosphere, \colorbox[RGB]{255,235,205}{but} \colorbox[RGB]{255,235,205}{worst} of all not GREAT beer! & neg./pos. \\
\bottomrule[1pt]
\end{tabular}}
\caption{The saliency map of two examples comparing Baseline and CRR models. \underline{Aspects} are underlined.}
\label{tab:saliency}
\end{table}

Finally, we use the gradient salience maps~\cite{DBLP:conf/iui/AlqaraawiSWCB20} to visualize what different models focus on. Specifically, for each instance, we calculate the gradient of the cross entropy loss with respect to the output hidden state of the last layer of the pre-training model encoder. We then give colors to the sentence tokens according to their gradient norms: the tokens with larger gradient norms are given darker colors, signifying that they are more important to the model. We remove \textless cls\textgreater, \textless sep\textgreater, punctuation, aspect words and consider only the remaining words. Table~\ref{tab:saliency} shows some of the instances where we see in some cases, CRR indeed improves the performance by focusing on the predictive features. For example, in the sentence “3d rendering not slows it down considerably.”, the CRR model focuses on “not slows” from which it deduces that the sentiment is positive. We also show cases where our model fails to capture correct features as shown in the second example of Table~\ref{tab:saliency}. In this case, we hypothesise that this could be due to the less frequent token `waitstaff' in the sentence. We are also aware that saliency may not be the optimal technique for visualising the subtle changes \cite{kim2019saliency}. We leave in depth interpretations for future work.

\section{Related Works}
\label{relatedwork}
Recent works in NLP focus on questions relating to the robustness of opaque neural network based models \cite{hsieh2019robustness} and many inspiring approaches have been proposed to improve model robustness Wang et al. \shortcite{wang2021measure}. Our work follows \cite{buhlmann2020invariance,DBLP:journals/corr/abs-2109-00725,veitch2021counterfactual,DBLP:conf/aaai/WangC21} that uses causal theory to study robustness issues. Recently, counterfactually-augmented data has shown their benefits on improving model robustness~\cite{DivyanshKaushik2019LearningTD,DBLP:conf/aaai/WangC21}. However, producing high-quality augmented data is expensive and time-consuming~\cite{PdraigCunningham2022ExploringTE}, and automatically generated methods are limited on the explicitly available causal structure; given these limitations, we have shown that counterfactual generation might in practice lead to only mediocre results. Our work inspired by Mitrovic et al. \shortcite{DBLP:conf/iclr/MitrovicMWBB21} focus on learning task specific invariant representations via non-counterfactual data augmentation.

Recent methods for improving robustness on ARTs have been approached in other ways as well, such as introducing position-bias \cite{DBLP:journals/corr/abs-2105-14210} or combining dependency paring trees \cite{DBLP:conf/naacl/HouQWYHHZ21}. We detail the methods in Section~\ref{models}. Our proposal instead uses a causal lens and exploits noisy augmented data. Our approach can be further combined with the above methods to further boost robustness performance on ARTs and other datasets~\cite{jiang-etal-2019-challenge}.

\section{Conclusions}
We propose an effective method to improve aspect robustness by exploiting non counterfactually-augmented data in a causal framework. Our proposal consists of adding noisy cost-effective data augmentation and relying on invariances to effectively promote causally relevant representations. Experimental results show that our method can improve over the strong baseline on both task-specific and robustness datasets. Our observations also suggest that the proposed method can improve performance with augmented data that is not aligned with test distribution and has been shown to transfer well across multiple domains.



\bibliography{anthology,ccl2023-en}

\newpage
\appendix
\section{Appendices}
The implementation of the framework, the models and the experimental logs will be made available post publication.

\subsection{KL Divergence During Training}
\begin{figure}[h]
    \centering
    \includegraphics[width=0.5\textwidth]{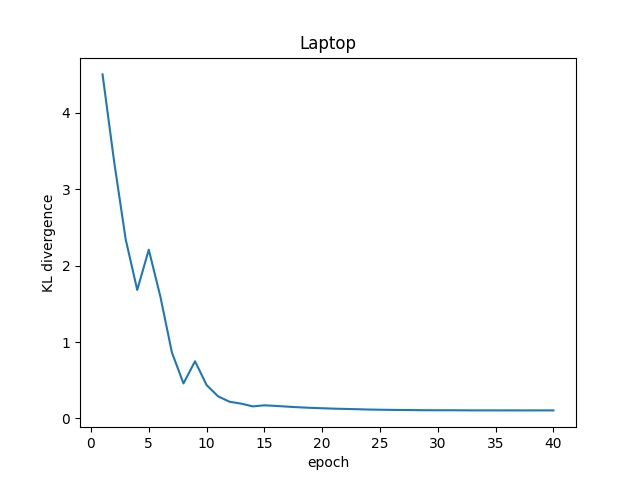}
    \includegraphics[width=0.5\textwidth]{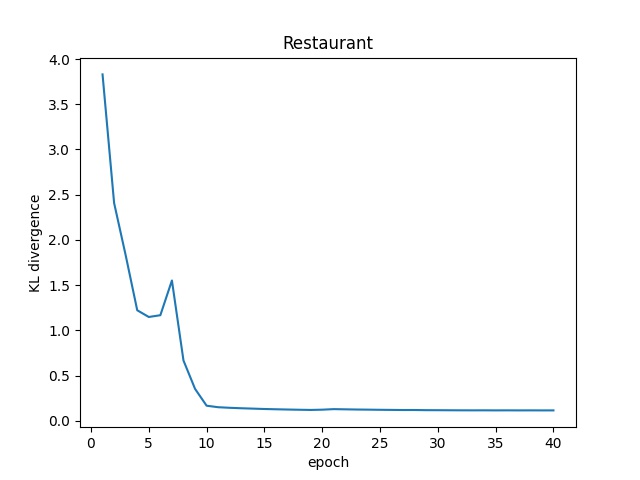}
    \caption{Trend of KL divergence while training our approach. We sum up the kl divergence value of all the instances in training set at the end of each epoch.}
    \label{fig:kl_trend}
\end{figure}

\paragraph{KL regularization term is truly optimized during training.} For RoBERTa${\rm _L}$+CRR models, we visualize the trend of KL divergence in Figure~\ref{fig:kl_trend}. The results show that the KL divergence is minimized through training despite some fluctuations in the first few epochs.

\paragraph{Results of models with KL divergence in the other direction and JS divergence.} We also conduct experiments on RoBERTa${\rm_L}$ using other divergence regular terms. Specifically, we replace the Equation~\ref{kl} with the following equations:
\begin{equation}
\nonumber
\begin{aligned}
\mathcal{L}_{KL}&=\infdiv{p(y\,\vert\,x^a)}{p(y\,\vert\,x)}\quad\forall x,y\in\mathcal{D}\\
\mathcal{L}_{JS}&=\jsdiv{p(y\,\vert\,x)}{p(y\,\vert\,x^a)}\quad\forall x,y\in\mathcal{D}
\end{aligned}
\end{equation}
and the rest remain the same. As the results shown in Table~\ref{tab:divergence}, we find that although the method was still effective, the improvement on ARS is not as significant as the direction in Equation~\ref{kl}.

\begin{table}[h]\footnotesize
\centering 
\resizebox{0.5\textwidth}{!}{
\begin{tabular}{lccc}
\toprule[1pt]
\multirow{2}{*}{Method} 
  & 
  \multicolumn{3}{c}{\textsc{Restaurant}} \\
  \cmidrule(lr){2-4}
  &
  Acc.(O) & Acc.(N) & ARS(N) \\
\midrule[0.5pt]

RoBERTa${\rm_L}$
& 86.73 & 81.32 & 59.48 \\
\midrule[0.5pt]
+our method
& \textbf{87.63} & \textbf{82.79} & \textbf{63.30}\\
+KL another direction
& 87.04 & 82.14 & 61.19 \\
+JS divergence
& 87.41 & 82.78 & 62.82 \\

\bottomrule[1pt]
\end{tabular}}
\caption{The accuracy of models with other divergence on SemEval2014 Restaurant (Acc.(O)) and ARTs (Acc.(N)). The models are based on RoBERTa${\rm_L}$ pretrained model. For ARTs's Restaurant, we also show ARS (ARS(N)).}
\label{tab:divergence}
\end{table}

\begin{figure}[h]
\centering
\begin{tikzpicture}
\pgfplotsset{every axis legend/.append style={at={(0.47,0.34)},
anchor=north west},
every axis y label/.append style={at={(0.07,0.5)}}}
\begin{axis}[xlabel=Regularization Weights,
    ylabel=Accuracy (\%),xtick=data,legend columns=2,
    legend style={font=\tiny},font=\footnotesize,
    width=8cm,height=3.5cm,
    ymin=70,ymax=85,ytick={70,75,80,85}]
\addplot table [x=a, y=Laptop-acc, col sep=comma] {db1.csv};
\addplot table [x=a, y=Restaurant-acc, col sep=comma] {db1.csv};
\legend{Laptop, Restaurant}
\end{axis}
\end{tikzpicture}

\caption{Accuracy (\%) of the RoBERTa${\rm _L}$+CRR models on ARTs. The models are trained with the same hyperparameters except for the regularization weights.}
\label{fig:com_weights}
\end{figure}
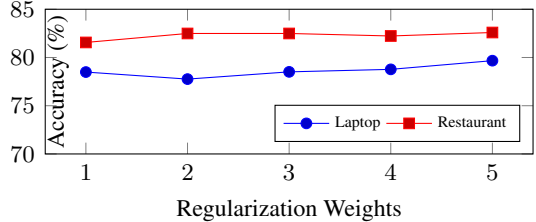

\paragraph{Our approach is insensitive to the regularization weight.} We conduct experiments over different regularization weights $\{1,2,3,4,5\}$ for the same model with the same hyperparameters. The result in Figure~\ref{fig:com_weights} shows that different weights result in quite similar improvements in the model performance.

\subsection{CRR and Adversarial results with \textsc{AddDiff}}
\label{appendix:AddDiff}
We experiment on the original \textsc{AddDiff} proposed by Xing et al. \shortcite{DBLP:conf/emnlp/XingJJWZH20} to further show the models' performance when using different augmented data (i.e. \textsc{AddDiff} and \textsc{AddDiffMix}). Specifically, we first apply \textsc{AddDiff} operation to the training set in SemEval2014 Restaurant without manual adjustment, then train the models by adversarial training (RoBERTa${\rm_L}$+Adv/\textsc{AddDiff}) and CRR (RoBERTa${\rm_L}$+CRR/\textsc{AddDiff}) respectively. We present the results in Table~\ref{tab:adv_nomix_res}.

Compared with RoBERTa${\rm_L}$+Adv using \textsc{AddDiffMix}, we find that the model performance deteriorates by about 0.63\% in accuracy for the original test set and 0.44\% in accuracy (0.77\% in ARS) when using \textsc{AddDiff}; this might be due to that the model learns the wrong bias to ignore the text near the end. However, RoBERTa${\rm_L}$+CRR performs quite similarly when using these two versions of augmented data. In both cases, CRR outperforms both the RoBERTa${\rm_L}$ and the adversarial training baselines.

\begin{table}[h]\footnotesize
\centering 
\resizebox{0.6\textwidth}{!}{
\begin{tabular}{lcccc}
\toprule[1pt]
\multirow{2}{*}{Model} &
\multirow{2}{*}{Aug.} &
  \multicolumn{3}{c}{\textsc{Restaurant}} \\
  \cmidrule(lr){3-5}
  & &
  Acc.(O) & Acc.(N) & ARS(N) \\
\midrule[0.5pt]

RoBERTa${\rm_L}$ & --
& 86.73 & 81.32 & 59.48 \\
\midrule[0.5pt]

\multirow{2}{*}{RoBERTa${\rm_L}$+Adv} &
    \textsc{AddDiff} & 86.23 & 81.51 & 58.50 \\
& 
    \textsc{AddDiffMix} & 86.86 & 81.95 & 59.27 \\
\midrule[0.5pt]

\multirow{2}{*}{RoBERTa${\rm_L}$+CRR} &
    \textsc{AddDiff} & 87.59 & \textbf{82.97} & 63.05 \\
& 
    \textsc{AddDiffMix} & \textbf{87.63} & 82.79 & \textbf{63.30} \\

\bottomrule[1pt]
\end{tabular}}
\caption{Model performance comparison using \textsc{AddDiff} and \textsc{AddDiffMix} data augmentations, respectively.}
\label{tab:adv_nomix_res}
\end{table}

\subsection{Proof Of Theorem}
\label{appendix:causal theorem}
To mathematically define what it means for task $Y^R$ cover the causal relationships in task set $\mathcal{Y}$, we first need to introduce relationship cover from the perspective of graph theory.

\noindent
\textbf{Definition 1. (Relationship Cover).} \emph{Let $\sim$ and $\approx$ be two edges sets from the directed graph $\mathcal{G}$. If any subset of $\sim$ is a subset of $\approx$, then we call $\approx$ cover the edges in $\sim$.}

\noindent
Furthermore, if the causal graph of a specific task $Y^R$ \emph{relationship cover} the causal graph of every $Y_t$ in task set $\mathcal{Y}$, we say $Y^R$ cover all the relationships in $\mathcal{Y}$.

\noindent
\textbf{Definition 2. (Invariant Representation).} \emph{Let $X$ and $Y$ be the input and target, respectively. We call $R(X)$ an invariant representation for $Y$ under spurious pattern $S$ if
\begin{equation}
P(Y\,\vert\,R(X))=P(Y\,\vert\,do(s_i),R(X))\quad\forall s_i\in\mathcal{S},\notag
\end{equation} 
where $do(s_i)$ denotes assigning spurious pattern $S$ the value $s_i$ and $\mathcal{S}$ is the domain of $S$.
}

\noindent
\textbf{Theorem.} \emph{Let $\mathcal{Y}=\{Y_t\}_{t=1}^T$ be a family of tasks. Let $Y^R$ be a specific task cover all the causal relationships in $\mathcal{Y}$. If $R(X)$ is an invariant representation for $Y^R$ under spurious pattern interventions $S$, then $R(X)$ is an invariant representation for all tasks in $\mathcal{Y}$ under spurious pattern interventions $S$, i.e. 
\begin{equation}
\begin{aligned}
P(Y^R\,\vert\,R(X))&=P(Y^R\,\vert\,do(s_i),R(X))\Rightarrow\\
P(Y_t\,\vert\,R(X))&=P(Y_t\,\vert\,do(s_i),R(X))\quad\forall s_i\in\mathcal{S},Y_t\in\mathcal{Y}\notag
\end{aligned}
\end{equation} 
Thus $R(X)$ is a representation that generalizes to $\mathcal{Y}$.} 

\noindent
\textbf{Proof.} Let $t\in\{1,...,T\}$. We have
\begin{equation}
\begin{aligned}
P(Y_t\,\vert\,R(X))&=\int\,P(Y_t\,\vert\,Y^R)P(Y^R\,\vert\,R(X))dY^R\\
&=\int\,P(Y_t\,\vert\,Y^R)P(Y^R\,\vert\,do(s_i),R(X))dY^R\\
&=P(Y_t\,\vert\,do(s_i),R(X)).\notag
\end{aligned}
\end{equation}

For the last equality, we use the mechanism that $Y_t\,\vert\,Y^R$ is independent of $S$, i.e. $P(Y_t\,\vert\,Y^R)=P(Y_t\,\vert\,do(s_i),Y^R)$. The second equality follows the assumption that $R(X)$ is an invariant representation for $Y^R$ under changes in $S$. Therefore we get that $R(X)$ is an invariant representation for all $Y_t$ under changes in $S$.

\end{document}